\documentclass{article}

\usepackage{microtype}
\usepackage{graphicx}
\usepackage{subfigure}
\usepackage{booktabs} 
\usepackage{amssymb}
\usepackage{amsmath}
\usepackage{amsthm}
\usepackage{amsfonts}
\usepackage{dsfont}
\usepackage[utf8]{inputenc} 
\usepackage[T1]{fontenc}

\usepackage{bbm}
\usepackage{multirow}

\usepackage{tabularx}
\usepackage{arydshln}
\usepackage[hang,flushmargin]{footmisc}

\newcommand{\ignorethis}[1]{}

\newtheorem*{conjecture*}{Conjecture}

\newcommand{\xpar}[1]{\noindent\textbf{#1}\ \ }

\newcommand{\comm}[1]{}

\newcommand\blfootnote[1]{%
  \begingroup
  \renewcommand\thefootnote{}\footnote{#1}%
  \addtocounter{footnote}{-1}%
  \endgroup
}

\makeatletter
\def\blfootnote{\gdef\@thefnmark{}\@footnotetext}
\makeatother

\usepackage{hyperref}
\usepackage{caption}
\usepackage{xurl}

\usepackage[accepted]{icml2021}

\icmltitlerunning{ Hierarchical Relationship Alignment Metric Learning }

\begin{document}

\twocolumn[
\icmltitle{Hierarchical Relationship Alignment Metric Learning}

\icmlsetsymbol{equal}{*}

\begin{icmlauthorlist}
\icmlauthor{Lifeng Gu}{tju}
\end{icmlauthorlist}

\icmlaffiliation{tju}{Tian Jin university}

\icmlcorrespondingauthor{}{gulifeng666@163.com}
\icmlkeywords{Machine Learning, ICML}

\vskip 0.3in
]



\begin{abstract}
Most existing metric learning methods focus on learning a similarity or distance measure relying on similar and dissimilar relations between sample pairs.
  However, pairs of samples cannot be simply identified as similar or dissimilar in many real-world applications, e.g., multi-label
  learning, label distribution learning.
  To this end, relation alignment metric learning (RAML)  framework is proposed to handle the metric learning problem in those scenarios. But RAML learn a linear metric, which can't model complex datasets. Combining with deep learning and RAML framework, 
we propose a hierarchical relationship alignment metric leaning model HRAML, which uses the concept of relationship alignment to model metric learning problems under multiple learning tasks, and makes full use of the consistency between the sample pair relationship in the feature space and the sample pair relationship in the label space. Further we organize several experiment divided by learning tasks, and verified the better performance of HRAML against many popular methods and RAML framework.
\end{abstract}

\section{Introduction}
In many computer vision and pattern recognition tasks, e.g., face recognition \cite{guillaumin2009you}, image classification \cite{mensink2012metric}, and person re-identification \cite{liao2015person}, it is crucial to learn a discriminative distance metric to measure the similarity between pairs of samples.
However, for some learning tasks, e.g., multi-label learning \cite{zhang2015lift} and label distribution learning \cite{geng2016label}, relations between sample pairs cannot be simply identified as similar or dissimilar. Thus, the existing metric learning methods cannot work on the above tasks.
The problem arises that it is difficult to classify two images into similar or dissimilar sample pair.
Above discussions encourage us to propose a generalized metric learning method, which can be flexibly adopted to various kinds of tasks. RAML~\cite{zhu2018} framework was proposed to handle the problem.
Combing with deep neural networks and RAML, we propose a hierarchical relationship alignment metric leaning model HRAML, which uses the concept of relationship alignment to model metric learning problems under multiple learning tasks, and makes full use of the consistency between the sample pair relationship in the feature space and the sample pair relationship in the label space. 
\blfootnote{\noindent \textbf{Affiliations} \hfill $^{1}$Tju}
\blfootnote{\textbf{Correspondence} \hfill Lifeng Gu - \texttt{gulifeng666@163.com}}

\section{Related works}
\xpar{deep metric learning }
deep metric learning  want to learn a good metric to 
measure the similarity of samples. Contrative loss~\cite{hadsell2006dimensionality} was used to combine with Siamese network and achieved good results. The loss of two-tuples can be used to model several problems. It is also a simple and feasible method. The sampling complexity is $o(n^{ 2})$, easier than many subsequent variants. ~\citet{wen2016discriminative} combines softmax and center loss for face recognition. Center loss is used to minimize the distance from the sample representation to the center of the category. ~\citet{ge2018deep} combined the triple loss and hierarchical tree to propose a new loss function. He used the hierarchical tree to encode the context of the sample, and then used the hierarchical tree to calculate the intra-class distance and the inter-class distance. Sample the more suitable triples, and finally use the dynamic interval mechanism to form the loss function. ~\citet{wang2017deep} proposed a novel angle loss to improve the discriminative ability of sample representation. Different from the two-tuple and triple-tuple loss, the angle loss uses the angle of the sample to the center of the category to impose constraints. The angle constraint has rotation and scaling invariance. It naturally encodes the geometric relationship between the three samples. You can set the angle The size to change the constraint strength has the advantage of being very robust. The slow convergence speed of binary or ternary loss is mainly due to the fact that only one negative sample is compared in each update and other negative samples are ignored. Npair loss~\cite{sohn2016improved} improves this point, and compares batches in each update All other negative samples in each update. Circle loss~\cite{sun2020circle} was proposed, Circle loss re-weights the similarity between samples to highlight the incomplete sample similarity. Circle loss unifies two basic learning methods in the field of feature learning: learning based on category labels and learning based on pair label. \citet{cakir2019deep} proposed FastAP, which learns features by optimizing the average accuracy of the ranked list. FastAP connects feature learning and ranking problems together, opening up new ideas.

\section{Background}
 ~\citet{zhu2018} proposed the concept of relational alignment in the field of metric learning. We briefly describes it as follows:
For metric learning, pairwise constraints are often used to describe the sample pair relationship in the decision space.
Here, we introduce relation alignment learning to metric learning.
 $f({\bf{x}}_i, {\bf{x}}_j, {\bf{M}},b)$ is used to measure the sample relationship in the feature space, $g(\bf {y_i},\bf {y_j})$ are used to measure the sample relationship in the decision space.
$g(\bf{ y_i},\bf {y_j})$ is specifically designed to handle different tasks.
In addition, $\bf{A} \in {\mathbb{R}}^{n \times n}$ and $\bf{E} \in {\mathbb{R}}^{n \times n}$ are used as The sample relationship matrix between the feature space and the decision space, the relationship alignment means
The sample relationship of the feature space should be consistent with the sample relationship of the decision space. This is a wide-ranging idea,
\begin{eqnarray}
\small
\left[ {\begin{array}{*{20}{c}}
{{a_{11}}}&{...}&{{a_{i1}}}&{...}&{{a_{n1}}}\\
{...}&{...}&{...}&{...}&{...}\\
{{a_{1i}}}&{...}&{{a_{ii}}}&{...}&{{a_{ni}}}\\
{...}&{...}&{...}&{...}&{...}\\
{{a_{1n}}}&{...}&{{a_{in}}}&{...}&{{a_{nn}}}
\end{array}} \right] = \left[ {\begin{array}{*{20}{c}}
{{e_{11}}}&{...}&{{e_{i1}}}&{...}&{{e_{n1}}}\\
{...}&{...}&{...}&{...}&{...}\\
{{e_{1i}}}&{...}&{{e_{ii}}}&{...}&{{e_{ni}}}\\
{...}&{...}&{...}&{...}&{...}\\
{{e_{1n}}}&{...}&{{e_{in}}}&{...}&{{e_{nn}}}
\end{array}} \right]
\nonumber
\end{eqnarray}
 $a_{ij}$ and $e_{ij}$ represent the relationship between the sample ${\bf{x}}_i$ and ${\bf{x}}_j$ in the feature space and the decision space.
 In order to maintain consistency, here is a requirement
\begin{equation}
    f({\bf{x}}_i, {\bf{x}}_j, {\bf{M}},b)=g({\bf{y}}_i,{\bf{y}}_j ).
\label{consistency}
\end{equation}
$g(\bf{y_i},\bf{y_j})$ represents the different degrees of the two samples in the decision space.
$g({\bf{y}}_i,{\bf{y}}_j)$ reflects the relationship between sample pairs in the decision space and is used to guide the  learning in the feature space $({\bf{M}},b )$ .
\begin{equation}
\begin{array}{l}
f({{\bf{x}}_i},{{\bf{x}}_j},{\bf{M}},b) = {\left( {{{\bf{x}}_i} -{{\bf{x}}_j}} \right)^T}{\bf{M}}\left( {{{\bf{x}}_i}-{{\bf{x}}_j} } \right) + b\\
{\rm{ }} \qquad \qquad \qquad \; = \left\langle {{\bf{M}},{{\bf{T}}_{ij}}} \right\rangle + b
\end{array}
\end{equation}
$\left\langle {\cdot ,\cdot} \right\rangle $ is defined as the inner product of two Frobenius matrices, $b$ is the offset, and ${{\bf{T}}_{ij}} = \left( {{{\bf{x}}_i}-{{\bf{x}}_j}} \right){\left( {{{\bf{x}}_i}-{{\bf {x}}_j}} \right)^T}$.
 \eqref{consistency} can be rewritten as
\begin{equation}
    g({{\bf{y}}_i},{{\bf{y}}_j}) = \left\langle {{\bf{M}},{{\bf{T}}_{ij} }} \right\rangle + b
    \label{linearregression}
\end{equation}
Once the relational function $g(\bf{y_i},\bf{y_j})$ is selected, the formula \eqref{linearregression} can be regarded as a linear regression problem.
Here, the metric learning problem is transformed into solving a sample pair regression problem, and the input is a sample pair $({\bf{x}}_i,{\bf{x}}_j)$.

\section{HRAML}
 Given sample pairs ${\bf{x}}_i$ and ${\bf{x}}_j$, after being modeled by a neural network of $M+1$ layer, they can finally be formalized as $f(\bf {x_i})={\bf{h}}_i^{(M)}=s( {\bf{W}}^{{(m)}^{T}}{\bf{x}}_i ^{(m-1)}+{\bf{b}}^{(m)})$ and
$f({\bf x_j})={\bf{h}}_j^{(M)}=s( {\bf{W}}^{{(m)}^{T}}{\bf{x} }_{j}^{(m-1)}+{\bf{b}}^{(m)})$, where the mapping $s: {\mathbb{R}}\to {\mathbb{R} } $ is a nonlinear activation function, and $m$ is the number of layers. ${\bf{x_i}}^{(m)} = {\bf{W}}^{{(m)}^{T}}{\bf{x}}_{i}^{(m-1) }+{\bf{b}}^{(m)}$.\par
Given a sample pair $x_{i}$ and $x_j$, using the idea of relationship alignment, the alignment relationship between the sample pair in the feature space and the decision space can be formalized as:
\begin{equation}
R(f(x_i),f(x_j)) = g(y_i,y_j)
\label{aligment}
\end{equation}
Where $R$ represents the sample pair relation function in the feature space, and $g$ represents the relation function in the feature space. By specifying the relationship functions $R$ and $g$, different relationships specifically used for alignment in the idea of relationship alignment can be expressed.
We defines the sample pair relationship of the feature space as the distance between them: $R(f(x_i),f(x_j))=D_{ij}=\left\|f({\bf x_i})-f({ \bf x_j})\right\|_2$, and then use the mean square error loss optimize formula \eqref{aligment} to derive the objective function of HRAML:
\begin{equation}
\mathop {\min} J=
\frac{1}{4}\sum\nolimits_{(i,j)}^n
(D_{ij}^2-g(y_i,y_j))^2+r(\theta)
\label{hraml}
\end{equation}
 $r(\theta)=\lambda\sum_{m=1}^M \left(\left\|{\bf{W}}^{(m)}\right\|_F^2+\left\|{\bf{b}}^{(m)}\right\|_2^2\right)$ is a regularization item, $g(y_i,y_j)$ is relation function.
In order to optimize  \eqref{hraml}, we use stochastical gradient descent and we have flowing equations:\\
for layer $m = M$ in neural network:
\begin{equation}
\begin{split}
\frac{\partial J(i,j)}{\partial {\bf{W}}^{(m)}}& = 
\left(D^2_{ij}-g(y_i,y_j)\right)\Big( \frac{\partial s({\bf{x}}_i^{(m)})}{\partial {\bf{x}}_i^{(m)}}\left ({f({\bf{x}}_i)}-{f({\bf{x}}_j)}\right)\\
&{\bf{h}}_i^{{(m-1)}^{T} }
-\frac{\partial s({\bf{x}}_j^{(m)})}{\partial {\bf{x}}_j^{(m)}}\left ({f({\bf{x}}_i)}-{f({\bf{x}}_j})\right){\bf{h}}_j^{{(m-1)}^{T}}  \Big )\\
&={\bf{ \delta}}_i^{(m)}
{\bf{h}}_i^{{(m-1)}^{T}}
-{\bf{\delta}}_j^{(m)}{\bf{h}}_j^{{(m-1)}^{T}}
\label{j/wm}
\end{split}
\end{equation}
\begin{equation}
\begin{split}
\frac{\partial J(i,j)}{\partial {\bf{b}}^{(m)}}& =
    \left(D^2_{ij}-g(y_i,y_j)\right) \Big( \frac{\partial {s({\bf{x}}_i^{(m)})}}{\partial {\bf{x}}_i^{(m)}}
-\frac{\partial{s({\bf{x}}_j^{(m)}})}{\partial {\bf{x}}_j^{(m)}}  \Big )
\\&\left(f({\bf{x}}_i)-{f({\bf{x}}_j)}\right)
= {\bf{\delta}}_i^{(m)}-{\bf{\delta}}_j^{(m)}
\label{j/bm}
\end{split}
\end{equation}
for layer $m = 1,2\dots M-1$ in neural network:
\begin{equation}
\begin{split}
\frac{\partial J(i,j)}{\partial {\bf{W}}^{(m)}} &=\frac{\partial {s({\bf{x}}_i^{(m)}})}{\partial {\bf{x}}_i^{(m)}} {{\bf{W}}^{(m+1)}}^{T}\frac{\partial J(i,j)}{\partial{{\bf{{x}}_i}^{(m+1)}}}{{\bf{h}}_i^{(m-1)}}^{T}\\
&- \frac{\partial {s({\bf{x}}_j^{(m)}})}{\partial {\bf{x}}_j^{(m)}}{{\bf{W}}^{(m+1)}}^{T}\frac{\partial J(i,j)}{\partial{{\bf{x}}_j}^{(m+1)}}{{\bf{h}}_j^{(m-1)}}^{T}\\
&={ \bf{\delta}}_i^{(m)} {{\bf{{h}}}_i^{(m-1)}}^{T}-{\bf{\delta}}_j^{(m)}{{{\bf{h}}_j^{(m-1)}}}^{T}
\end{split}
\label{j/wm1}
\end{equation}
\begin{equation}
\begin{split}
\frac{\partial J(i,j)}{\partial {\bf{b}}^{(m)}} &=\frac{\partial{ s({\bf{x}}_i^{(m)}})}{\partial {\bf{x}}_i^{(m)}} {\bf{W}}^{{(m+1)}^{T}}\frac{\partial J(i,j)}{{\partial{\bf{x}}_i}^{(m+1)}}- \frac{\partial {s({\bf{x}}_j^{(m)}})}{\partial {\bf{x}}_j^{(m)}}\\
&{{\bf{W}}^{{{(m+1)}}^{T}}}\frac{\partial J(i,j)}{\partial{{({{\bf{x}}_j}^{(m+1)})}}}
={ \bf{\delta}}_i^{(m)}-{\bf{\delta}}_j^{(m)}
\label{j/bm1}
\end{split}
\end{equation}
The corresponding metric learning HRAML algorithm is summarized in Algorithm \ref{alg:Hraml}.
\begin{algorithm}[t]
\caption{The algorithms of our proposed HRAML}
\begin{algorithmic}[1]
\REQUIRE ~~\\
{Training data $\bf{X} \in {\mathbb{R}}^{d \times m}$, where $d$ and $m$ are the numbers of feature dimension and samples, respectively. number of
network layers $M$, learning
rate $\mu$, iterative number $T$}

\STATE Generate sample pairs $({\bf{x}}_{i1},{\bf{x}}_{i2})$, $i=1,2,...,n$.\\
\STATE Compute sample relation
$g({\bf{x}}_{i1},{\bf{x}}_{i2})$, $i=1,2,...,n$.\\
\REPEAT{ 
\STATE {Randomly select a pair$(\bf{x}_{i}, \bf{x}_{j})$ from sample  pairs.}
\STATE 
Set ${\bf{h}}_{i}^{0}$= ${\bf{x}}_{i}$, ${\bf{h}}_{j}^{0}$ = ${\bf{x}}_{j}$.\\
    \FOR{m = 1,2,\dots,M}
  { \STATE Do forward propagation to get ${\bf{h}}^{m}_{i}$ and ${\bf{h}}^{m}_{j}$.}
   \ENDFOR
 
   \FOR{m = M,M-1,\dots,1}
  { \STATE Do back propagation to get $\frac{\partial J(i,j)}{\partial {\bf{W}}^{(m)}}$ and $\frac{\partial J(i,j)}{\partial {\bf{b}}^{(m)}}$ by \eqref{j/wm},\eqref{j/bm},\eqref{j/wm1},\eqref{j/bm1}}
\ENDFOR
\STATE
{update ${\{{\bf{W}}^{m}, {\bf{b}}^{m}\}}, m=1,2,\dots,M$ by gradient descent.}}
 \UNTIL{converge}
\ENSURE ~~\\
  {Weights and biases: \{{{\bf{W}}$^{m}$}, {\bf{b}}$^{m}$\}, m=1,2,\dots,M}
\end{algorithmic}
\label{alg:Hraml}
\end{algorithm}
\section{Implementation Details}
\subsection{Training Details}
Due to the diversification of the training data that needs to be processed,we uses general mlp as the encoder and tanh as the activation function.
 In addition, we normalizes the output of the network so that a better performance can be obtained~\cite{wu2017sampling}, the normalized form is
\begin{equation}
    f(x) = \frac{f(x)}{\|f(x)\|}
\end{equation}
 f(x) is the output of the network. \\
 We initializes ${\bf{b}}^{m}, m=1,2,\dots, M$ is 0, and ${\bf{W}}^{m}, m=1,2,\ dots,M$ is initialized to the distribution $U[-0.2, 0.2]$.
 \subsection{Relation function}
 In the feature space, we need to use differentiable functions as the relation function to ensure the stability and ease of learning. We uses ordinary Euclidean distance as the relation function. In the decision space, we can imitate the feature space and use the function with parameters, which will make HRAML method has a broader form and can be linked to more methods, but for the convenience of the sample selection stage, we are consistent with~\cite{zhu2018}, using a fixed function without parameters as the decision space Function, we consider the $l_1$ norm:
 Let ${\bf{y}}_i$ and  ${\bf{y}}_j$ represent label of ${\bf{x}}_i$ and ${\bf{x}}_j$.
Relation function will be :
\begin{equation}
    g({{\bf{y}}_i},{{\bf{y}}_j}){\rm{  = }}{\left\| {{{\bf{y}}_i} - {{\bf{y}}_j}} \right\|_1}
\label{relationg}
\end{equation}
 where ${\left\| {\bf{a}} \right\|_1}{\rm{ }}$ is $l_1$-norm of item $\bf{a}$.
\subsection{Sample selection problem}
In the process of neural network learning, sample selection is very important and has a great impact on the results. We can uses difficult sample mining \cite{shrivastava2016training}.
\section{Experiment}
In this section, we conduct experiments to validate the performance of the proposed HRAML.
We consider three applications, including single-label classification, multi-label classification, label distribution learning
. The following part will be organized as the corresponding parts.
 \subsection{Classification}
\noindent\textbf{Experiment setup.}
 Following \cite{zhu2018}, we use "S/F/C" represents the number of samples, features and classes.
We compare our method with  popular methods, e.g., ITML \cite{davis2007information}, LMNN \cite{weinberger2009distance}, DML \cite{ying2012distance}, DoubletSVM (DSVM) \cite{wang2015kernel},GMML \cite{zadeh2016geometric} and RAML\cite{zhu2018}. For fair comparison, the parameters of all compared methods are set as the default setting of the original
references.\\
\noindent\textbf{Experimental analysis.}
 Table \ref{table2} list the classification accuracy of  different metric learning methods on many datasets, respectively, where the best results are marked in bold face.
 HRAML get the best performance, it beat others methods, not only linear methods but also kernel methods based nolinear methods. HRAML has a clear and sample objective function Under classification task, neural network can optimise easily and learn discriminate metric. 
\begin{table*}[htbp!]
\centering
 \tabcolsep=0.1in
 \resizebox{\textwidth}{!}{
    \begin{tabular}{ccccccccccccc}
   \hline
    Data &S/F/C  & ITML  & LDML  & LMNN  & DSVM & GMML  & DML   & RAML-SVR   & RAML-KRR & HRAML \\
       \hline
    binalpha&1404/320/36  & 0.6303$ \pm $0.0501 & 0.6542$ \pm $0.0317 & 0.6112$ \pm $0.0358 & 0.5625$ \pm $0.0322 & 0.5338$ \pm $0.1986 & 0.5063$ \pm $0.0251 & \textbf{\underline{{0.7250$ \pm $0.0348}}} & \textbf{0.6850$ \pm $0.0351} &\textbf{0.7243$ \pm $0.0262} \\
    caltech101&8641/256/101  & \textbf{0.5803$ \pm $0.0162} & 0.5528$ \pm $0.0157 & 0.5795$ \pm $0.0126 & 0.5584$ \pm $0.0159 & 0.5500$ \pm $0.0117 & 0.3936$ \pm $0.0123 & \textbf{{0.5855$ \pm $0.0095}} & \textbf{0.5803$ \pm $0.0147}& \textbf{\underline{{0.6015$ \pm $0.0161}}}\\
    MnistDat&3495/784/10  & 0.8695$ \pm $0.0142 & 0.8858$ \pm $0.0124 & 0.8721$ \pm $0.0255 & 0.8848$ \pm $0.0194 & 0.8589$ \pm $0.0171 & 0.8323$ \pm $0.0239 & \textbf{0.9019$ \pm $0.0175} & \textbf{\underline{0.9087$ \pm $0.0137}}& \textbf{{0.8983$ \pm $0.0168}}
    \\
    Mpeg7 &1400/6000/70  & 0.8214$ \pm $0.0333 & 0.7971$ \pm $0.0365 & 0.8253$ \pm $0.0232 & 0.8271$ \pm $0.0353 &\textbf{ 0.8429$ \pm $0.0228 }& 0.7071$ \pm $0.0267 & \textbf{\underline{{0.8450$ \pm $0.0305}}} & 0.7936$ \pm $0.0341 & 0.8043$ \pm $0.0248
    \\

    news20&3970/8014/4  & 0.8678$ \pm $0.0200 & 0.8816$ \pm $0.0145 & 0.8734$ \pm $0.0290 & 0.8594$ \pm $0.0159 & 0.8647$ \pm $0.0143 & 0.8166$ \pm $0.0222 & \textbf{0.9025$ \pm $0.0132} & \textbf{{0.9217$ \pm $0.0145}}
    & \textbf{\underline{0.9436$ \pm $0.0133}}\\
    TDT2\_20&1938/3677/20  & 0.9587$ \pm $0.0358 & 0.9531$ \pm $0.0306 & 0.9352$ \pm $0.0197 & 0.9499$ \pm $0.0175 & 0.9437$ \pm $0.0275 & 0.6333$ \pm $0.0176 & \textbf{0.9679$ \pm $0.0244} & \textbf{{0.9762$ \pm $0.0164}} & \textbf{\underline{{0.9860$ \pm $0.0154}}}
    \\
    uspst&2007/256/10  & 0.8979$ \pm $0.0261 & 0.9084$ \pm $0.0243 & 0.9096$ \pm $0.0217 & 0.9125$ \pm $0.0172 & 0.8858$ \pm $0.0168 & 0.8030$ \pm $0.0330 & \textbf{{0.9525$ \pm $0.0147}} & \textbf{0.9447$ \pm $0.0157}
    & \textbf{\underline{0.9537$ \pm $0.0202}}\\
  \hline

 \end{tabular}
 }
 \vspace{-0.01cm}

  \caption{classification accuracy on many datasets}
   \label{table2}
\end{table*}%
\begin{table}[H]\small
\centering
\tabcolsep=0.01in
\begin{tabular}{c|c|cccc}
\hline
&{ Data }& emotion & flags  & corel800 \\
\hline
\multirow{5}{*}{ MLKNN} &{Hamming Loss}$\downarrow$& 0.2137  & 0.3099    &  0.0137  \\
                        &{Ranking Loss}$\downarrow$& 0.1729  & 0.2228   &  0.1888  \\
                        &{One Error}$\downarrow$   & 0.3317  & 0.2154    &  0.6825  \\
                        &{Coverage}$\downarrow$    & 1.9158  & 3.8154    & 88.5100  \\
                        &{Average Precision}$\uparrow$ & 0.7808  & 0.8084    &  0.3276  \\
\hline

\multirow{5}{*}{ RAML-SVR}   &{Hamming Loss}$\downarrow$& \textbf{0.2054}  & \textbf{{0.2967}}    &  \textbf{0.0135}  \\
                        &{Ranking Loss}$\downarrow$& \textbf{0.1577} & \textbf{0.2179}    &  \textbf{{0.1882}}  \\
                        &{One Error}$\downarrow$   & \textbf{\underline{0.2376}}  & \textbf{{0.2000}}    &  \textbf{{0.6425}}  \\
                        &{Coverage}$\downarrow$    & \textbf{1.8960}  &\textbf{ 3.8115}    & \textbf{{88.2350}}  \\
                        &{Average Precision}$\uparrow$  & \textbf{0.8101}  & \textbf{{0.8128}}    &  \textbf{0.3386}  \\

\hline\multirow{5}{*}{ RAML-KRR}   &{Hamming Loss}$\downarrow$& \textbf{{0.2046}}  & \textbf{{0.2967}}    &  \textbf{\underline{0.0134}}  \\
                        &{Ranking Loss}$\downarrow$& \textbf{\underline{0.1382}} & \textbf{{0.2113}}    &  \textbf{0.1888}  \\
                        &{One Error}$\downarrow$   & \textbf{0.2574}  & \textbf{{0.2000}}    &  \textbf{0.6550}  \\
                        &{Coverage}$\downarrow$    & \textbf{{1.7327}}  &\textbf{ \underline{3.7692}}    & \textbf{88.5100}  \\
                        &{Average Precision}$\uparrow$  & \textbf{\underline{0.8225}}  & \textbf{0.8112}    &  \textbf{\underline{0.3388}}  \\

\hline
\multirow{5}{*}{ HRAML} &{Hamming Loss}$\downarrow$& 0.2060  & \textbf{\underline{0.2791 }}   &  \textbf{0.0135}  \\
                        &{Ranking Loss}$\downarrow$& 0.1690  & \textbf{\underline{0.2021}}   & \textbf{\underline{ 0.1861 }} \\
                        &{One Error}$\downarrow$   &\textbf{ 0.2871}  & \textbf{{0.2154}}    &  \textbf{\underline{0.6075}}  \\
                        &{Coverage}$\downarrow$    & 1.8861  & \textbf{\underline{3.6769}}    & \underline{87.6025}  \\
                       
                        &{Average Precision}$\uparrow$  & \textbf{0.8052}  & \textbf{\underline{0.8244}}    &  \textbf{\underline{0.3582}}  \\
                        \hline
\end{tabular}
\caption{ results on muti-label datasets}
\vspace{-0.3cm}
\label{mlknnresult}
\end{table}.
\begin{table}[H]\footnotesize
\centering
\tabcolsep=0.01in

\begin{tabular}{c|ccccccc}
\hline
Criterion &{Chebyshev}$\downarrow$                            & {Clark}$\downarrow$         &{Canberra}$\downarrow$    &{Cosine}$\uparrow$  & {Intersection}$\uparrow$    \\
\hline
  AAKNN   & 0.3261  & 1.8448 & 4.3412 & 0.6905 & 0.5506\\

 RAML-SVR& {\textbf{0.3102}}  & \textbf{1.6986} & \textbf{3.8576} & \textbf{0.7051}  & \textbf{0.5739}\\

 RAML-KRR&\textbf{{0.3139}}& \textbf{\underline{1.6865}} & \textbf{\underline{3.8419}}& \textbf{{0.7057}}& \textbf{{0.5743}}   \\

 HRAML&\textbf{\underline{0.2907}}&\textbf{1.7664} &\textbf{4.0922} &\textbf{\underline{0.7321}}&\textbf{\underline{0.5929}}\\

\hline
\end{tabular}
\caption{ results on Nature Scene dataset.}
\label{ldlrsult}
\end{table}
\subsection{Multi-label Learning}
\noindent\textbf{Dataset.}
Like \cite{zhu2018}, we use three multi-label datasets to evaluate performance\footnote{http://mulan.sourceforge.net/datasets-mlc.html}, emotion \cite{trohidis2008multi}, flags, and corel800 dataset \cite{hoi2006learning}.\par
\noindent\textbf{Evaluation Method.}
We use the performance of the MLKNN algorithm to evaluate the discriminative ability of the learned metric,
We use multiple popular multi-label learning indicators used in \cite{zhu2018} to evaluate the permormance of MLKNN. The up arrow represents the higher the indicator, the better, and the down arrow represents the lower the indicator, the better. \par
\noindent\textbf{Experimental Analysis.}
Table \ref{mlknnresult} shows the results on multi-label datasets,
Compared with the original performance of MLkNN, RAML and HRAML learn more discriminative metric in multi-learning task. Compared with RAML algorithm, HRAML has better experimental results. Thanks to the improvement of the encoder, it has better performance than RAML. Beyond linear transform or kernel methods, HRAML has more power to extract features. And  it can distinguish samples under different labels well through its optimization goals, .
\subsection{Label Distribution Learning}
\noindent\textbf{dataset.}
We use the Nature Scene dataset used in \cite{zhu2018}.\\
\noindent\textbf{Evaluation Method.}
We use the performance of AAKNN to evaluate the learned metric, and use AAKNN and RAML as comparison algorithms. We use  multiple  evaluation indicators used in \cite{zhu2018} as the evaluation indicators.
 In order to maintain consistency with other experimental parts, We use \eqref{relationg} as the relationship function of the decision space. Of course, some better distribution functions such as kl divergence can be used as the relationship function, which should achieve better effect.
 \noindent\textbf{Experimental analysis.}
Table \ref{ldlrsult} shows the comparison of AAKNN, RAML and HRAML.
Thanks to the HRAML's encoder and optimization goal, it's performance is better than RAML.

\section{Conclution}
This paper proposes a hierarchical relationship alignment model HRAML, which uses the concept of relationship alignment to model metric learning problems under multiple learning tasks, and makes full use of the consistency between the sample pair relationship in the feature space and the sample pair relationship in the label space. Finally, the performance of HRAML is verified under a variety of tasks.

\bibliography{bib}
\bibliographystyle{icml2021}

\end{document}